# Modified Topological Image Preprocessing for Skin Lesion Classifications

Hong Cheng, Rebekah Leamons, Ahmad Al Shami

Department of Computer Science

Southern Arkansas University - Magnolia, AR, USA

hcheng@saumag.edu

**Abstract:** This paper proposes a modified Topological Data Analysis model for skin images preprocessing and enhancements. The skin lesion dataset HAM10000 used with the intention of identifying the important objects in relevant regions of the images. In order to evaluate both the original dataset and the preprocessed dataset, Deep Convolutional Neural Network and Vision Transformer models were utilized to train both models. After training, the experimental results demonstrate that the images preprocessed using the Modified Topological Data Analysis consistently perform better.

## 1. Introduction

Compared to other countries, the US has the highest number of skin cancer cases annually [1, 2]. One in five Americans will develop skin cancer by age 79 [3]. Though there are treatment options, early detection is vital for effective treatment. Skin cancer can be tricky to diagnose because of the number of different skin lesions that may or may not be cancerous. This study focuses on diagnosing the following seven types of skin lesions:

1. Melanocytic Nevi
2. Melanoma
3. Benign Keratosis-like Lesions
4. Basal Cell Carcinoma
5. Actinic Keratoses
6. Vascular Lesions
7. Dermatofibroma

Of these, only Melanoma and Basal Cell Carcinoma are inherently cancerous.
More recently, an object detection method based on Topological Data Analysis (TDA) and consisting of Topological Image Modification (TIM) and Topological Image Processing (TIP) was proposed [4].

## 2. Proposed Framework

The primary goal of our research is to preprocess using an enhanced TDA model developed specifically to deal with classifying the HAM 10000 [6] skin lesion. The proposed models train both the original dataset and the topologically preprocessed dataset using a deep Convolutional Neural Network and Visual Transformer as our training models, and their respective results was compared. Fig.1 shows the full-proposed skin legion framework followed by explanations of the used models starting with the TDA of Persistent Homology

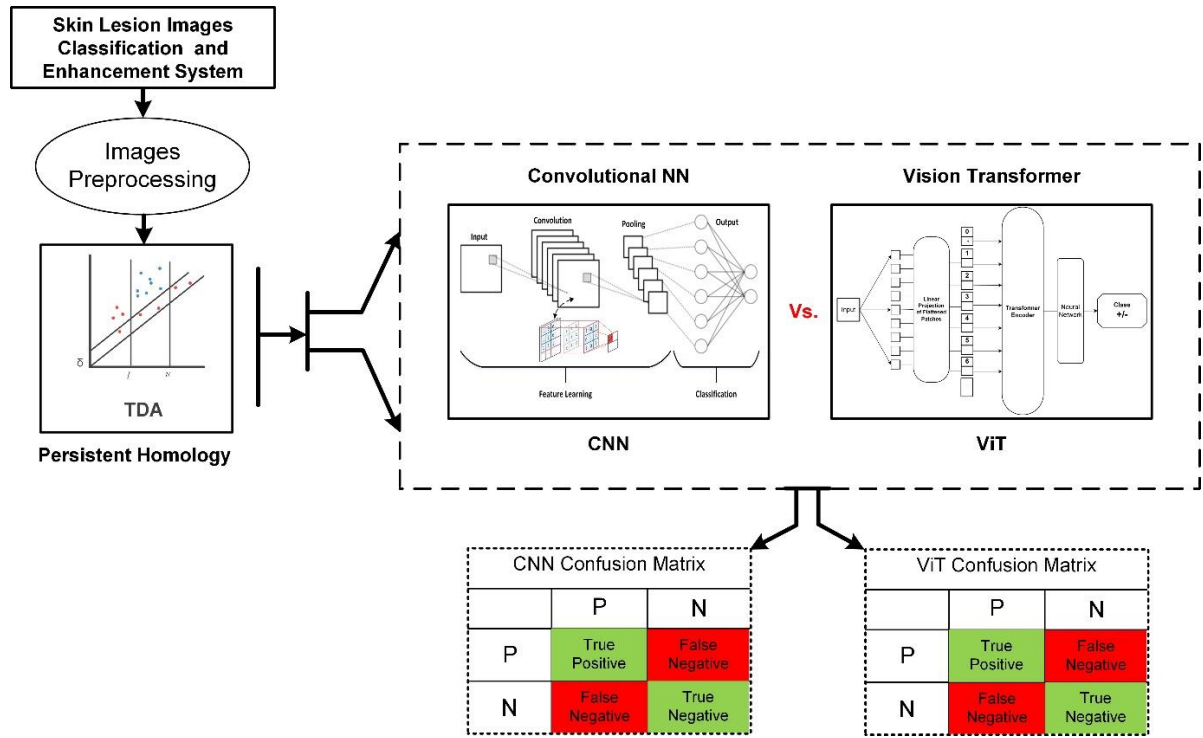

*Fig. 1: Illustrates the proposed framework for this study. The first stage is for the skin lesion image preprocessing using TDA of Persistent Homology, which is then pushed for classifications comparisons using Convolutional NN vs. Vision Transformer. Both classification performance results are then presented in the separate confusion matrix.*

**Persistent Homology of Images**
Persistent Homology [5] is a tool commonly used in Topological Data Analysis (TDA) [7]. It works by extracting features that represent the number of connected components, cycles, and voids. It also extracts these features' birth and death during the iterative process called filtration. Each of these features is summarized as a point in a persistence diagram. Persistent homology relies on objects referred to as simplices, which are the building blocks of simplicial complexes. A k-simplex is the convex hull of k+1 affinely independent points. A 0-simplex is also called a vertex, a 1-simplex an edge, and a 2-simplex a triangle.

The RGB image M has first converted into grayscale then the values of its pixels are used to define a filtration. The filtration $F$ for image M is defined as
$$K_i := (\{\sigma \in K: f(K) \leq i\})$$
Where
$$f: K \to R: \sigma \to \max_{p \in \sigma} gray_M(p)$$
And $K_0 \subseteq K_1 \subseteq K_2 ... \subseteq K_T = K$. The corresponding image $I_i$ as having a pixel value of 1 if the corresponding pixel is in $K_i$, and 0 otherwise. A connected component in $K_i, i \in \{1,2,...,T\}$, is then a maximal connected cluster of pixels with every value equal to 1.

An important rule to merge components used is called the elder rule which states that if two components or holes are merged when constructing the persistence diagram, the one with the greatest birth time dies.

**Topological Image Modification (TIM)**
*Image smoothing*
This process reduces the effect of outliers by defining a new image from the old image. This is done by averaging over the "neighborhood" of the old image to generate a new image.
*Border modification*
The idea of border modification is based on assumption that in many real-world images, the object(s) of interest do not connect to the border of the image, but the background does. Border modification constructs a new image by ensuring that every pixel within a certain distance of the border reaches the lowest value, while other values remain unchanged.

**Topological Image Processing (TIP)**
With the help of TIM, Topological image processing (TIP) is an algorithm to iterate over all lifetimes in decreasing order based on persistent diagrams and select threshold by taking the average between the two lifetimes where the largest of these differences is achieved. Then process images to increase the contrast between the objects with a lifetime above the threshold, and the rest (the background) of the image. Components are merged using the elder rule. Results include a binary image, marking objects of the original image consistent with (the number of) inferred components through its persistence diagram.

**Convolutional Neural Network**
A Convolutional Neural Network works to classify images in two phases: feature learning and classification. In feature learning, an input is first given to the program. In this case, input includes an image. Then, convolution layers are used to create a feature map. Convolution is an image processing technique that uses a weighted kernel (square matrix) to revolve over the image, multiply and add the kernel elements with image pixels. These convolution layers work by applying filters and padding to the input. These filters create a map of any important features on an image — in this study; these important features could be the shape or color of the skin lesion. Next, the pooling layer utilized dimensionality reduction techniques to reduce the size of the layer. Finally, the data is classified into a predetermined number of categories. This method is visualized below.

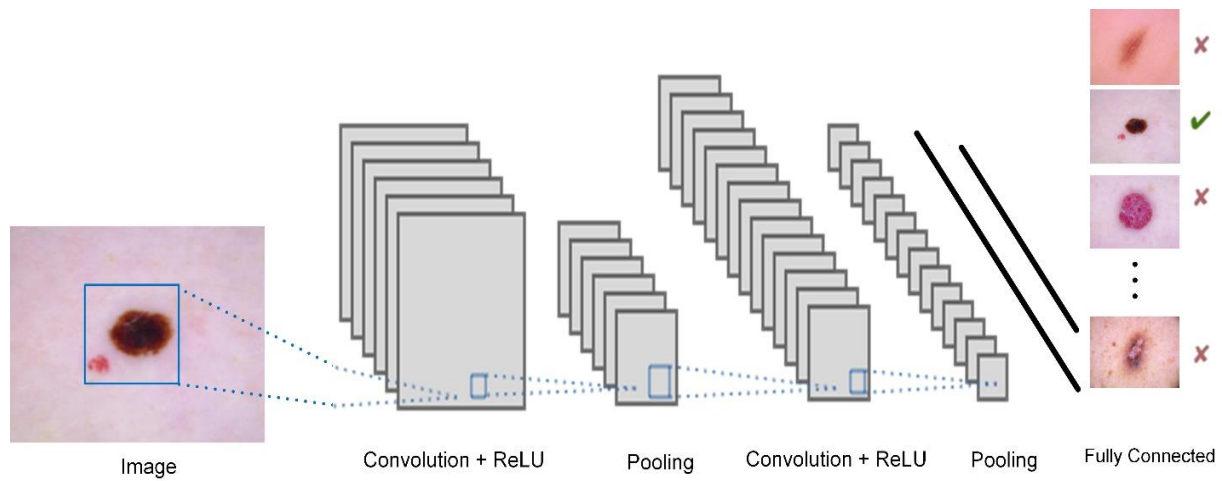

*Fig. 2: This diagram shows the phases of a CNN. This includes taking input, running it through a convolution layer, pooling, and classifying the data for output.*

**Vision Transformers**

Transformers are a relatively new tool in the realm of computer vision, but recent studies have proven their potential [9]. Transformers were used in language processing. These language Transformers would attempt to relate a word in a sentence to other words to find "context." In this way, we can consider each input image as a sentence that the Transformer is attempting to find the "context" of. In this case, the context refers to one of the seven possible skin lesion classifications. To do this, the Vision Transformer first splits an input image into patches, where every patch has a related position. These patches will then be run through encoders which try to relate one patch to another. Once the Transformer finds the "context," the image is run through a neural network and classified. A diagram of this process is shown below.

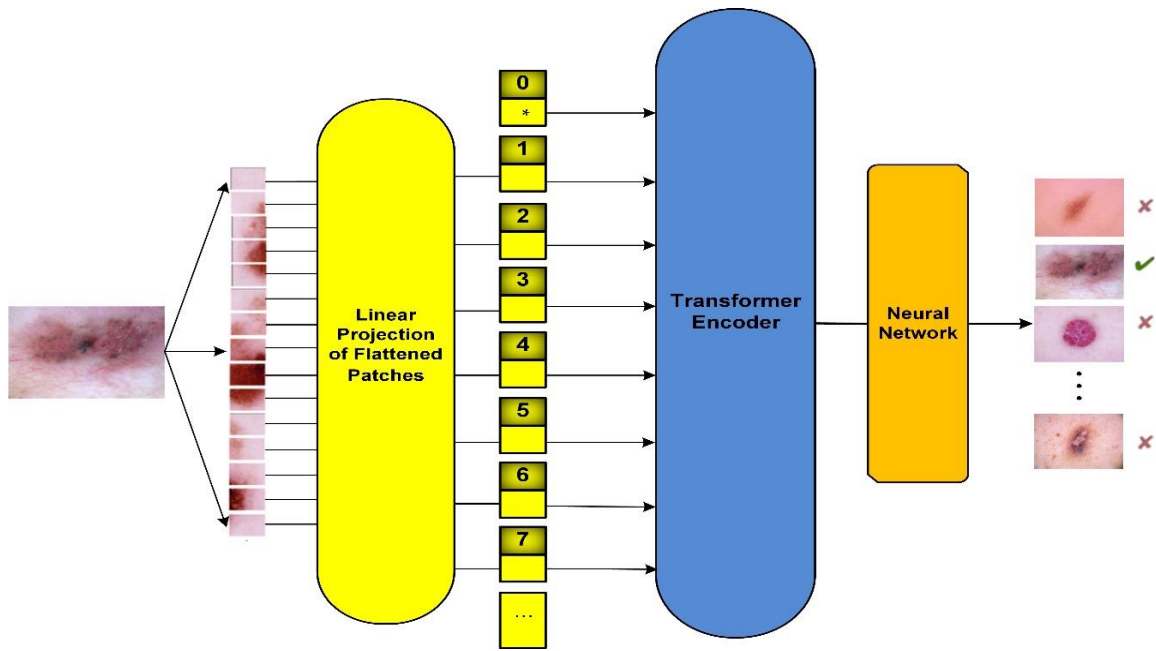

*Fig. 3. This diagram shows how a Vision Transformer works, using an example from the dataset. As seen in the diagram, the transformer takes an image, splits it into patches, creates a linear projection of these patches, assigns a location to the patches, runs them through an encoder, then sends the data through a neural network, before finally classifying the image.*

## 3. Methodology
The used dataset and the proposed modified TDA are described in this section.

### Dataset Description
The HAM10000 [6] dataset has two zip files: AM10000_images_part1.zip with 5,000 JPEG files and AM10000_images_part2.zip with 5,015 JPEG files. The 10,015 dermatoscopic images are in seven different classes as below:
8. Melanocytic Nevi - Abnormally dark birthmarks or moles that are noncancerous.
9. Melanoma - A type of cancer that begins in melanocytes, which are the cells that control skin pigments. Melanoma is usually characterized by dark skin lesions.
10. Benign Keratosis-like lesions - Noncancerous skin growths that are usually a waxy brown, black, or tan.
11. Basal cell carcinoma - A type of cancer that usually develops on areas of skin exposed to the sun.
12. Actinic Keratoses - Noncancerous, rough/scaly patch of skin caused by years of sun exposure.
13. Vascular lesions - Relatively common abnormalities of skin and underlying tissues. Commonly known as birthmarks.
14. Dermatofibroma - Common, benign fibrous nodules most commonly found on women.

**Modified TDA Preprocessing and Modifications**

To begin, four folders are created: two training folders and two testing folders. The images in both respective folders contain identical images, but one contains the topologically processed images. The dataset was split, with 90% given for training and 10% given for testing. This created the following four folders:
- test_dir (unprocessed testing images, 10% of the dataset)
- test_dir_TIP (processed testing images, 10% of the dataset)
- train_dir (unprocessed training images, 90% of the dataset)
- train_dir_TIP (processed training images, 90% of the dataset)

in which the first and second contain the same data and the third and fourth contain the same data.

Topological Image Processing (TIP) was used to isolate the skin lesion in each image and focus on the skin lesion. To do this, we first generated grayscale versions of the original images. The TIP program then isolated the skin lesion by removing the background "noise," which in this case is the skin itself. Once the lesion was isolated, the original image could be cropped so it only contained the lesion and less of the skin around it. The original TIP algorithm generates a binary matrix the same size as the original image. It marks 1 for each pixel that has features and marks 0 for each pixel that is considered noise. As a modification, we determine the bounding box that surrounds all the ones.

Let column and row as two lists which holds maximum in each column and row
We could determine the top bound using a loop to find the first row with any 1 in it
*t=0*
*while not row[t]:*
    *t+=1*
We find **bottom** bound $b$, **left** bound $l$ and **right** bound $r$ in similar ways. The final output then returns a modified cropped image like *imgs[t:b,l:r]* . The *image* refers to the original image. Fig. 4 below shows the proposed preprocess procedure.

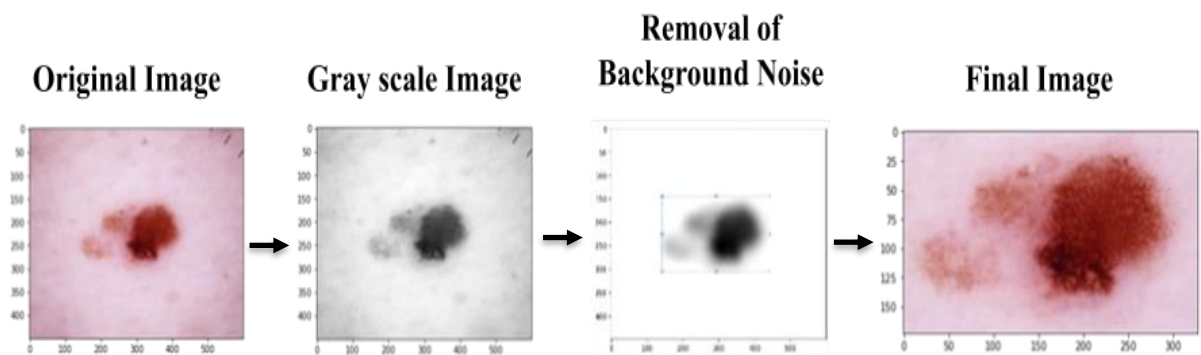

*Fig. 4: This diagram shows the modified TIP process by displaying the original image, the grayscale image, the removal of the background noise, and the final preprocessed image.*

These processed images were put into the appropriate test and training directories so the CNN and ViT models could be trained exclusively on the preprocessed and modified images.

## 5. Results and Analysis

The CNN trained on the base dataset had a test accuracy of 0.6816. When combined with topological data processing, the CNN had a test accuracy of 0.7136, showing a 3.20% improvement. The CNN vs. ViT with the modified TIP Confusion matrix results is presented in Fig. 5 below.

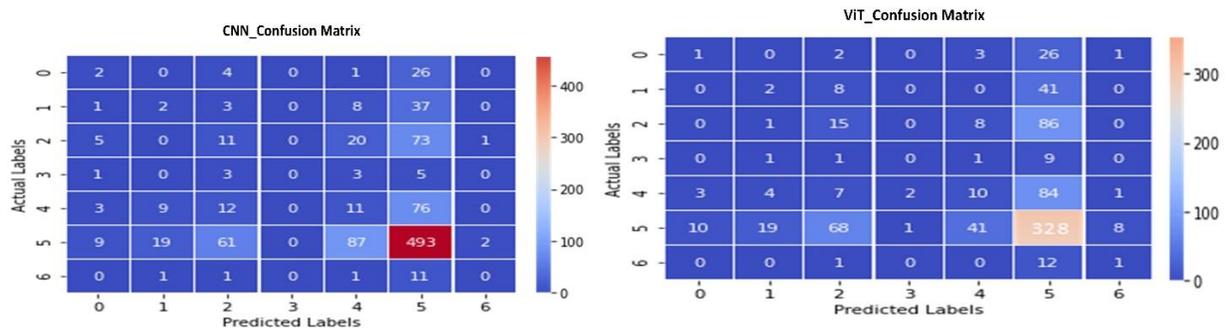

*Fig. 5.a (left) CNN with TIP Vs. Fig. 5b (right) ViT with TIP Confusion Matrices results.*

The graphs correlating to the CNN results are shown below:

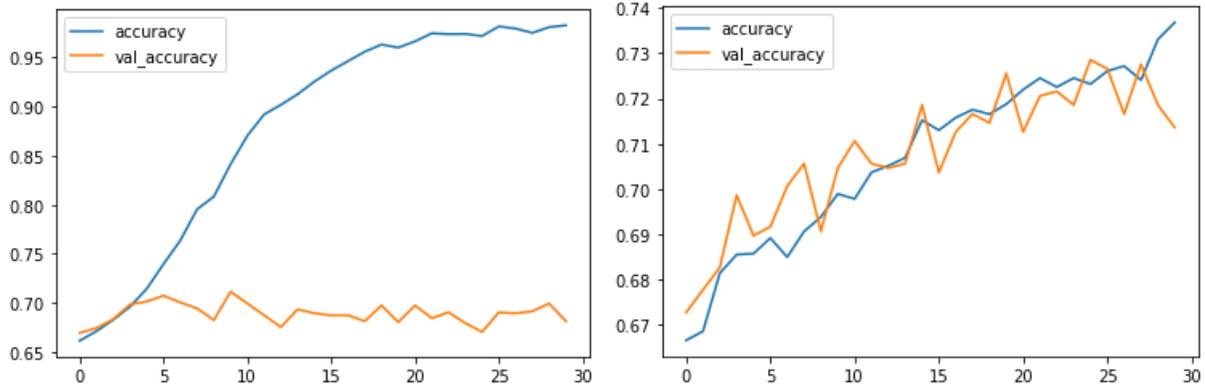

*Fig. 6.a (left): This graph shows the accuracy (blue) against the actual test accuracy (orange) of the CNN without TIM processing over the epochs, as seen on the x-axis. The test accuracy does not appear to improve with each epoch.*

*Fig. 6.b (right): This graph shows the accuracy (blue) against the actual test accuracy (orange) of the CNN with TIM processing over the epochs, as seen on the x-axis. The test accuracy is shown to mostly improve with each epoch.*

The Vision Transformer trained on the base dataset had a test accuracy of 0.8398. When combined with topological data processing, the Vision Transformer had a test accuracy of 0.8760, showing a 3.62% improvement. The graphs correlating to the Vision Transformer results are shown below:

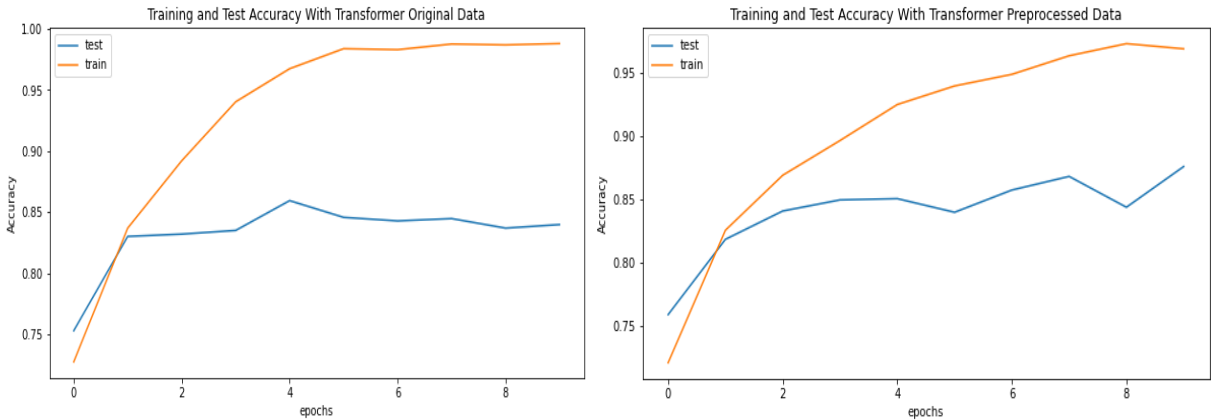

*Fig. 7.a (left): This graph shows the accuracy (blue) against the actual test accuracy (orange) of the Vision Transformer without TIM processing over the epochs, as seen on the x-axis. The test accuracy appears to plateau after its first epoch.*

*Fig. 7.b (right): This graph shows the accuracy (blue) against the actual test accuracy (orange) of the Vision Transformerwith TIM processing over the epochs, as seen on the x-axis. The test accuracy is shown to continue slightly increasing.*

## 6. Conclusion

Based on this study, we can make the reasonable conclusion that Topological Image Modification and Processing, when used to preprocess data, does indeed improve the accuracy of multiple deep learning models. Using this method, accuracies were increased by at least 3% in each tested model. This improved accuracy has exciting implications in many fields, including computer vision and medical diagnostics, as seen in this study.

In the future, we would like to compare the performance of topologically processed images versus unprocessed images using different deep learning models and methodologies, such as Residual Neural Networks or Soft Attention.

In addition, we hope to address the issue of overfitting. Overfitting occurs when a model fails to generalize and instead becomes too specific to the data at hand, which creates a high variance in model performance [14]. This issue occurred a few times throughout the process of this study, so we hope to isolate the cause and avoid this in the future.